\documentclass[letterpaper, 10 pt, conference]{ieeeconf}  

\IEEEoverridecommandlockouts                              
\usepackage[utf8]{inputenc}
\usepackage{amsmath, scalerel} 
\usepackage{amssymb}
\usepackage{amsopn}
\usepackage{graphicx}
\usepackage[style=base]{caption}
\usepackage{color}
\usepackage{quotmark}
\usepackage{booktabs}
\usepackage{todonotes}
\usepackage{lipsum}
\usepackage{import}
\usepackage{footnote}
\usepackage{multirow}
\usepackage{hhline}
\usepackage{subcaption}
\usepackage[pdfborder={0 0 0}, bookmarks=false, final]{hyperref}

\usepackage{floatrow}

\usepackage[abbreviations,binary-units]{siunitx}
\DeclareSIUnit{\mAh}{mAh}
\DeclareSIUnit{\Wh}{Wh}

\usepackage{todonotes}
\usepackage[gernotes]{optional}

\usepackage{scrextend}
\addtokomafont{labelinglabel}{\sffamily}

\usepackage{tikz}
\usetikzlibrary{calc}
\usetikzlibrary{matrix}
\usetikzlibrary{math}
\usetikzlibrary{plotmarks}
\usetikzlibrary{positioning}
\usetikzlibrary{shapes}
\usetikzlibrary{arrows}
\usetikzlibrary{arrows.meta}

\usepackage{pgfplots}
\pgfplotsset{compat=newest}
\usepgfplotslibrary{patchplots}
\usepackage{grffile}
\usepackage{ textcomp }

\usepackage{pgfplots}
\pgfplotsset{compat=newest} 
\pgfplotsset{plot coordinates/math parser=false} 
\newlength\figureheight 
\newlength\figurewidth 

\title{\LARGE \bf
Coordinated Heterogeneous Distributed Perception based on Latent Space Representation
}

\author{Timo Korthals$^{\dagger *}$, Jürgen Leitner$^{\text{\textdaggerdbl}}$ and Ulrich Rückert$^{\dagger}$
\thanks{
	$^{\dagger}$Bielefeld University,
	Cognitronics \& Sensor Systems,
	Inspiration 1, 33619 Bielefeld, Germany\newline
	\indent {$^\text{\textdaggerdbl}$}Australian Centre for Robotic Vision,
	Queensland University of Technology,
	Brisbane, Australia, 4000\newline
	\indent{$^{*}${\tt\small tkorthals@cit-ec.uni-bielefeld.de}}\newline
	\indent{}This research was supported by 'CITEC' (EXC 277) at Bielefeld University and the Federal Ministry of Education and Research (57388272). The responsibility for the content of this publication lies with the author.}
}

\begin{document}

\maketitle
\thispagestyle{empty}
\pagestyle{empty}

\begin{abstract}

We investigate a reinforcement approach for distributed sensing based on the latent space derived from multi-modal deep generative models.
Our contribution provides insights to the following benefits:
Detections can be exchanged effectively between robots equipped with uni-modal sensors due to a shared latent representation of information that is trained by a \textit{Variational Auto Encoder} (VAE).
Sensor-fusion can be applied asynchronously due to the generative feature of the VAE.
\textit{Deep Q-Networks} (DQNs) are trained to minimize uncertainty in latent space by coordinating robots to a Point-of-Interest (PoI) where their sensor modality can provide beneficial information about the PoI.
Additionally, we show that the decrease in uncertainty can be defined as the direct reward signal for training the DQN.

\end{abstract}

\section{INTRODUCTION \& RELATED WORK}

\textit{Active sensing} (AS) is one of the most fundamental problems and challenges in mobile robotics which seeks to maximize the efficiency of an estimation task by actively controlling the sensing parameters.
It can roughly be divided into two sub task:
the identification of a PoI (e.g. object or place) for exploration or to answer the question \textit{where to go next?} and the ability of a robot to navigate through the environment to reach a certain goal location without colliding with any obstacles en route.
Of particular interest for heterogeneous robot teams is the case where AS is used for determining the locations at which the robots should move to in order to acquire the most informative measurements.

The recent literature has seen a growing number of methods being proposed to tackle the task of autonomous navigation with \textit{deep reinforcement learning} (DRL) algorithms.
The works comprising \cite{Tai2016} formulate the navigation problem as \textit{Markov decision processes} (MDPs) or \textit{partially observable MDPs} (POMDPs) that first take in the sensor readings (color/depth images, laser scans, etc.) as observations and stack or augment them into states, and then search for the optimal policy by means of a \textit{deep neuronal network} (DNN) that is capable of guiding the agent to navigate to goal locations.
The first approach that performed well on such a generic task definition was DQN \cite{Mnih2013}.
\textit{Multi-agent reinforcement learning} (MARL) is the integration of \textit{multi-agent} systems with \textit{reinforcement learning} (RL), thus it is hence at the intersection of game theory and RL communities \cite{Nowe2012}.
However, while the automated discovery of the structure in raw data by a DNN is a time-consuming and error-prone task, an additional pre-processing step that performs feature extraction on the data helps to make DQNs more robust, and to downsize and even transfer them more easily.
\textit{Auto Encoder} (AE), \textit{Variational Auto Encoder} (VAE), and more recently \textit{Disentangled Variational Auto Encoder} ($\beta$VAE) have a considerable impact on the field of DRL as they encode the data into latent space features that are (ideally) linearly separable \cite{Higgins2017}.

\begin{figure*}
	\footnotesize
\begingroup
  \makeatletter
  \providecommand\color[2][]{
    \errmessage{(Inkscape) Color is used for the text in Inkscape, but the package 'color.sty' is not loaded}
    \renewcommand\color[2][]{}
  }
  \providecommand\transparent[1]{
    \errmessage{(Inkscape) Transparency is used (non-zero) for the text in Inkscape, but the package 'transparent.sty' is not loaded}
    \renewcommand\transparent[1]{}
  }
  \providecommand\rotatebox[2]{#2}
  \newcommand*\fsize{\dimexpr\f@size pt\relax}
  \newcommand*\lineheight[1]{\fontsize{\fsize}{#1\fsize}\selectfont}
  \ifx\svgwidth\undefined
    \setlength{\unitlength}{482.74548628bp}
    \ifx\svgscale\undefined
      \relax
    \else
      \setlength{\unitlength}{\unitlength * \real{\svgscale}}
    \fi
  \else
    \setlength{\unitlength}{\svgwidth}
  \fi
  \global\let\svgwidth\undefined
  \global\let\svgscale\undefined
  \makeatother
  \begin{picture}(1,0.23232863)
    \lineheight{1}
    \setlength\tabcolsep{0pt}
    \put(0,0){\includegraphics[width=\unitlength,page=1]{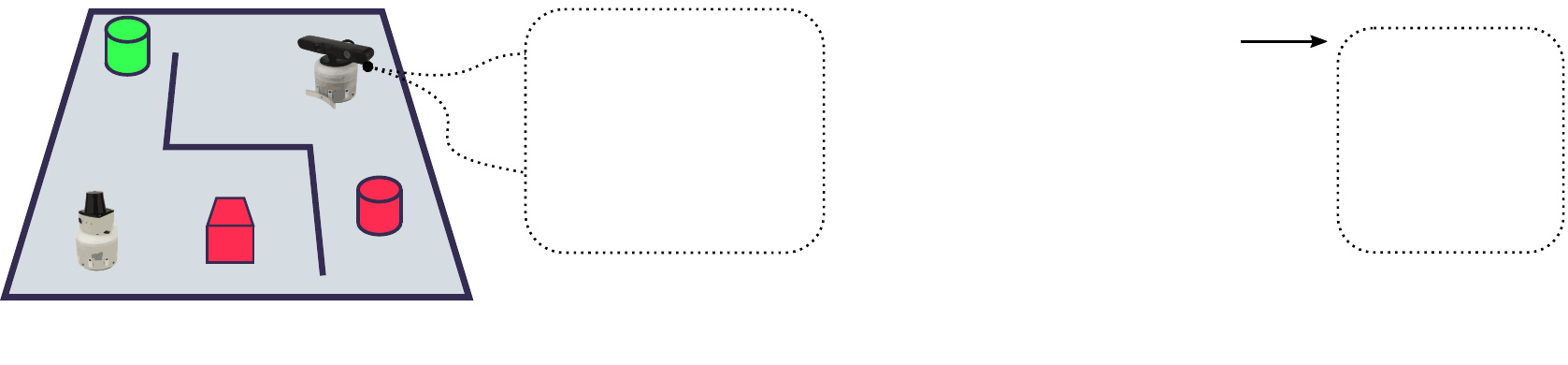}}
    \put(0.35839543,0.21798326){\color[rgb]{0,0,0}\makebox(0,0)[lt]{\begin{minipage}{0.2823508\unitlength}\raggedright $(x)$ or $(w)$ or $(x,w)$\end{minipage}}}
    \put(0.38479068,0.09732308){\color[rgb]{0,0,0}\makebox(0,0)[lt]{\begin{minipage}{0.13563046\unitlength}\raggedright $(x)$ and $(w)$\end{minipage}}}
    \put(0.43874704,0.15665702){\color[rgb]{0,0,0}\makebox(0,0)[lt]{\begin{minipage}{0.13563046\unitlength}\centering $(z)$\end{minipage}}}
    \put(0,0){\includegraphics[width=\unitlength,page=2]{architecture_RL_v3.pdf}}
    \put(0.66030277,0.09334923){\color[rgb]{0,0,0}\makebox(0,0)[lt]{\begin{minipage}{0.0653365\unitlength}\centering $D_z$\end{minipage}}}
    \put(0,0){\includegraphics[width=\unitlength,page=3]{architecture_RL_v3.pdf}}
    \put(0.71042861,0.2343998){\color[rgb]{0,0,0}\makebox(0,0)[lt]{\begin{minipage}{0.2182153\unitlength}\centering \textit{reward}\end{minipage}}}
    \put(0.82078808,0.11136438){\color[rgb]{0,0,0}\makebox(0,0)[lt]{\begin{minipage}{0.2182153\unitlength}\centering \textit{action}\end{minipage}}}
    \put(0,0){\includegraphics[width=\unitlength,page=4]{architecture_RL_v3.pdf}}
    \put(0.68091199,0.22197914){\color[rgb]{0,0,0}\makebox(0,0)[lt]{\begin{minipage}{0.13563046\unitlength}\centering $(z)$\end{minipage}}}
    \put(0.61375368,0.06475594){\color[rgb]{0,0,0}\makebox(0,0)[lt]{\begin{minipage}{0.13563046\unitlength}\centering mapping\end{minipage}}}
    \put(0,0){\includegraphics[width=\unitlength,page=5]{architecture_RL_v3.pdf}}
    \put(0.80324939,0.18831044){\color[rgb]{0,0,0}\makebox(0,0)[lt]{\begin{minipage}{0.2182153\unitlength}\raggedright \textit{state}\end{minipage}}}
    \put(0.30328064,0.02746921){\color[rgb]{0,0,0}\makebox(0,0)[lt]{\begin{minipage}{0.46655017\unitlength}\centering \textit{move\_base} by ROS for low-level control\end{minipage}}}
    \put(0,0){\includegraphics[width=\unitlength,page=6]{architecture_RL_v3.pdf}}
    \put(0.7695011,0.06380111){\color[rgb]{0,0,0}\makebox(0,0)[lt]{\begin{minipage}{0.17437156\unitlength}\centering \#modalities DQN\end{minipage}}}
    \put(0,0){\includegraphics[width=\unitlength,page=7]{architecture_RL_v3.pdf}}
    \put(0.36363868,0.06475594){\color[rgb]{0,0,0}\makebox(0,0)[lt]{\begin{minipage}{0.13563046\unitlength}\centering JMVAE\end{minipage}}}
    \put(0,0){\includegraphics[width=\unitlength,page=8]{architecture_RL_v3.pdf}}
  \end{picture}
\endgroup

	\caption{Architecture overview for the experiment. Multiple robots explore PoIs in the environment and encode the detected object by a VAE where its latent sample $z$ is mapped onto a environment representation shared by all robots. The current state is then used to derive a new policy to let the robots drive to objects where their modality can help to reduce uncertainty.}
	\label{fig:architecture_network}
\end{figure*}

The objective of this work is to investigate a reinforcement approach for distributed sensing based on the latent space derived from multi-modal deep generative models, as depicted in \autoref{fig:architecture_network}.
The main objective is to train multi-modal VAE that integrates all the information on different sensor modalities into a joint latent representation and then to generate one sensor information from the corresponding other one via this joint representation.
Therefore, this model can exchange multiple sensor modalities bi-directionally, for example, features from laser scanner data to images and vice versa, and can learn a shared latent space distribution between uni- and multi-modal cases.
Furthermore, we train a deep Q-Network that controls robots equipped with uni-modal sensors directly on the latent space with the objective of reducing uncertainty regarding detected objects. Our approach performs better than naive multi-robot exploration.
Our contribution makes use of the fields of deep neuronal networks for feature extraction, deep generative models for latent representations, and deep Q-Networks for optimal control in heterogeneous multi-agent systems in order to archive sufficient classification of objects in the environment.
Since this contribution concentrates on the generative models as the central feature enabling our approach, we stress this part in \autoref{seq:models}.
\autoref{seq:RL} provides an overview of our application within RL, which is then evaluated in \autoref{seq:setup}.

\section{GENERATIVE MODELS}
\label{seq:models}
\begin{figure}
	\footnotesize
\begingroup
  \makeatletter
  \providecommand\color[2][]{
    \errmessage{(Inkscape) Color is used for the text in Inkscape, but the package 'color.sty' is not loaded}
    \renewcommand\color[2][]{}
  }
  \providecommand\transparent[1]{
    \errmessage{(Inkscape) Transparency is used (non-zero) for the text in Inkscape, but the package 'transparent.sty' is not loaded}
    \renewcommand\transparent[1]{}
  }
  \providecommand\rotatebox[2]{#2}
  \newcommand*\fsize{\dimexpr\f@size pt\relax}
  \newcommand*\lineheight[1]{\fontsize{\fsize}{#1\fsize}\selectfont}
  \ifx\svgwidth\undefined
    \setlength{\unitlength}{226.77165354bp}
    \ifx\svgscale\undefined
      \relax
    \else
      \setlength{\unitlength}{\unitlength * \real{\svgscale}}
    \fi
  \else
    \setlength{\unitlength}{\svgwidth}
  \fi
  \global\let\svgwidth\undefined
  \global\let\svgscale\undefined
  \makeatother
  \begin{picture}(1,0.38204086)
    \lineheight{1}
    \setlength\tabcolsep{0pt}
    \put(0,0){\includegraphics[width=\unitlength,page=1]{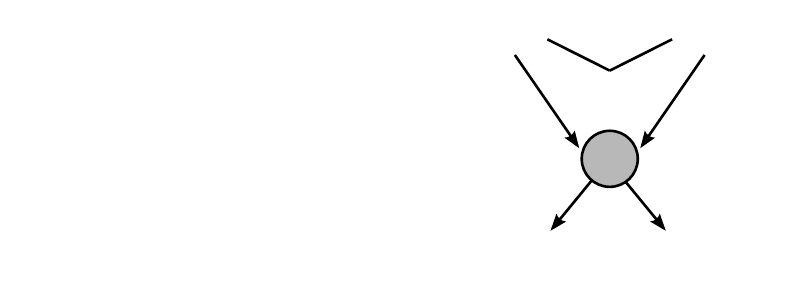}}
    \put(0.76210951,0.18985163){\color[rgb]{0,0,0}\makebox(0,0)[lt]{\begin{minipage}{0.61609064\unitlength}\raggedright $z$\end{minipage}}}
    \put(0,0){\includegraphics[width=\unitlength,page=2]{VAEs.pdf}}
    \put(0.66188813,0.3449835){\color[rgb]{0,0,0}\makebox(0,0)[lt]{\begin{minipage}{0.61609064\unitlength}\raggedright $x$\end{minipage}}}
    \put(0,0){\includegraphics[width=\unitlength,page=3]{VAEs.pdf}}
    \put(0.86872572,0.34498357){\color[rgb]{0,0,0}\makebox(0,0)[lt]{\begin{minipage}{0.61609064\unitlength}\raggedright $w$\end{minipage}}}
    \put(0,0){\includegraphics[width=\unitlength,page=4]{VAEs.pdf}}
    \put(0.68791307,0.06300513){\color[rgb]{0,0,0}\makebox(0,0)[lt]{\begin{minipage}{0.61609064\unitlength}\raggedright $x'$\end{minipage}}}
    \put(0,0){\includegraphics[width=\unitlength,page=5]{VAEs.pdf}}
    \put(0.83521939,0.0621385){\color[rgb]{0,0,0}\makebox(0,0)[lt]{\begin{minipage}{0.61609064\unitlength}\raggedright $w'$\end{minipage}}}
    \put(0.75463603,0.33434805){\color[rgb]{0,0,0}\makebox(0,0)[lt]{\lineheight{1.25}\smash{\begin{tabular}[t]{l}$q_{\phi}$\end{tabular}}}}
    \put(0,0){\includegraphics[width=\unitlength,page=6]{VAEs.pdf}}
    \put(0.63730727,0.22512373){\color[rgb]{0,0,0}\makebox(0,0)[lt]{\lineheight{1.25}\smash{\begin{tabular}[t]{l}$q_{\phi_{x}}$\end{tabular}}}}
    \put(0.86224793,0.22512373){\color[rgb]{0,0,0}\makebox(0,0)[lt]{\lineheight{1.25}\smash{\begin{tabular}[t]{l}$q_{\phi_{w}}$\end{tabular}}}}
    \put(0.63621272,0.13690409){\color[rgb]{0,0,0}\makebox(0,0)[lt]{\lineheight{1.25}\smash{\begin{tabular}[t]{l}$p_{\theta_{x}}$\end{tabular}}}}
    \put(0.8611534,0.13690409){\color[rgb]{0,0,0}\makebox(0,0)[lt]{\lineheight{1.25}\smash{\begin{tabular}[t]{l}$p_{\theta_{w}}$\end{tabular}}}}
    \put(0,0){\includegraphics[width=\unitlength,page=7]{VAEs.pdf}}
    \put(0.08128223,0.3449835){\color[rgb]{0,0,0}\makebox(0,0)[lt]{\begin{minipage}{0.61609064\unitlength}\raggedright $x$\end{minipage}}}
    \put(0,0){\includegraphics[width=\unitlength,page=8]{VAEs.pdf}}
    \put(0.07883652,0.18985162){\color[rgb]{0,0,0}\makebox(0,0)[lt]{\begin{minipage}{0.61609064\unitlength}\raggedright $z$\end{minipage}}}
    \put(0,0){\includegraphics[width=\unitlength,page=9]{VAEs.pdf}}
    \put(0.08128222,0.06300513){\color[rgb]{0,0,0}\makebox(0,0)[lt]{\begin{minipage}{0.61609064\unitlength}\raggedright $x'$\end{minipage}}}
    \put(0,0){\includegraphics[width=\unitlength,page=10]{VAEs.pdf}}
    \put(0.38280139,0.18985163){\color[rgb]{0,0,0}\makebox(0,0)[lt]{\begin{minipage}{0.61609064\unitlength}\raggedright $z$\end{minipage}}}
    \put(0,0){\includegraphics[width=\unitlength,page=11]{VAEs.pdf}}
    \put(0.30860496,0.06300513){\color[rgb]{0,0,0}\makebox(0,0)[lt]{\begin{minipage}{0.61609064\unitlength}\raggedright $x'$\end{minipage}}}
    \put(0,0){\includegraphics[width=\unitlength,page=12]{VAEs.pdf}}
    \put(0.45591128,0.0621385){\color[rgb]{0,0,0}\makebox(0,0)[lt]{\begin{minipage}{0.61609064\unitlength}\raggedright $w'$\end{minipage}}}
    \put(0,0){\includegraphics[width=\unitlength,page=13]{VAEs.pdf}}
    \put(0.30860487,0.3449835){\color[rgb]{0,0,0}\makebox(0,0)[lt]{\begin{minipage}{0.61609064\unitlength}\raggedright $x$\end{minipage}}}
    \put(0,0){\includegraphics[width=\unitlength,page=14]{VAEs.pdf}}
    \put(0.45591137,0.34498357){\color[rgb]{0,0,0}\makebox(0,0)[lt]{\begin{minipage}{0.61609064\unitlength}\raggedright $w$\end{minipage}}}
  \end{picture}
\endgroup

	\caption{Uni- and multi-modal VAE}
	\label{fig:VAE}
\end{figure}

The VAE proposed by \cite{Kingma2014} is used in settings where only a single modality $x$ is present, in order to find a latent encoding $z$ (c.f. \autoref{fig:VAE} right).
On the other hand, when multiple modalities $x$ and $w$ are available as shown in \autoref{fig:VAE} (mid.), it is less clear how to use the model as one would need to train a VAE for each modality.
Therefore, Suzuki et al. \cite{Suzuki2017} propose a joint multi-modal VAE (JMVAE) that is trained on two uni-modal encoder and bi-modal en-/decoders which share an objective function derived from the \textit{variation of information} (VI) between $x$ and $w$.
Therefore, the uni-modal encoders are trained, so that their distribution $q_{\phi_{x}}$ and $q_{\phi_{w}}$ are close to an encoder $q_{\phi}$ in order to build a coherent latent space posterior distribution.

With the JMVAE, we can extract joint latent features by sampling from the joint encoder $q(z|x,w)$ at testing time.
While the objective of \cite{Suzuki2017} is to exchange modalities bi-directionally ($x\rightarrow w'$ and $w\rightarrow x'$), our primary concern is to find a meaningful posterior distribution and hence, we analyze their approach with a view of using all statistics provided by the encoder network.

\section{APPLICATION TO RL}
\label{seq:RL}

One trend in current RL approaches is to first learn interpretable and factorized representations of the sensor data, rather than directly learn internal representations from raw observations \cite{Higgins2017}. 
Thus, (V)AEs are used to project high dimensional input data into a representation in which either a single or a group of latent units are sensitive to changes in single ground truth factors.
We follow this approach with the intention of using all the statistics advertised by an encoder network to derive reasonable navigation goals in a multi-agent setup.

The goal of an agent is to select navigation goals in the environment by selecting actions in a manner that lowers the variance of observations or vice versa, increasing the amount of information about a PoI $n$.
We define the properties of an MDP as follows:
$\mathcal{M}=\left( z_1, \ldots, z_N \right)$ is a global map of $N$ objects with feature $z$.
Each object $n$ can be observed by taking one action $a_n$, or the game can be terminated before observing all PoIs by the selection of \textit{no operation} (NOP): $\mathcal{A}=\left( a_1, \ldots, a_N, \text{NOP} \right)$, while dependent on the modality, an action $a_n$ samples from the posterior $q_{\phi_{x}}(z_n | x_n )$ or $q_{\phi_{w}}(z_n | w_n )$.
However, if there is a former observation by each other's modality, the encoder $q_{\phi}(z_n | x_n, w_n )$ is applied to perform sensor-fusion in place by generating the missing modality using $p_{\theta_{x}}(z_n)$ or $p_{\theta_{w}}(z_n)$ (c.f. \autoref{fig:sensor_fusion}).
Furthermore, the reward is defined by the increase in information $I$ after observation of object $n$ by $r \propto \Delta I_n=I^*_n - I_n$.

We apply a deep Q-Network to derive the next action $a_t$ given the current map $\mathcal{M}_t$.
For each agent, or more precisely each modality, an own Q-Network is trained.
It is worth mentioning that there is no interaction between the agents and thus, the learning process is decoupled since all robots act independently of one another.

\section{EXPERIMENTAL SETUP \& EVALUATION}
\label{seq:setup}

We performed experiments using two AMiRo mini-robots equipped with a camera and LiDAR \cite{Herbrechtsmeier2016a} to classify PoIs in the environment.
The overall mapping and decision architecture is depicted in \autoref{fig:architecture_network}.
Every robot explores the environment for PoIs and encodes the detection to a common map that is shared among all entities.
Based on this environmental model, a modality specific DQN decides which robot has to pursue which PoI to increase the information in the map.
It is worth mentioning, that the DQN only selects the target PoI and has no spatial information nor direct control at the motor level.
We use the DQN algorithm by \cite{Mnih2015} for training the deep Q-Network and sample the test and training data from the Gazebo simulator.

\begin{figure}
	\footnotesize
\begingroup
  \makeatletter
  \providecommand\color[2][]{
    \errmessage{(Inkscape) Color is used for the text in Inkscape, but the package 'color.sty' is not loaded}
    \renewcommand\color[2][]{}
  }
  \providecommand\transparent[1]{
    \errmessage{(Inkscape) Transparency is used (non-zero) for the text in Inkscape, but the package 'transparent.sty' is not loaded}
    \renewcommand\transparent[1]{}
  }
  \providecommand\rotatebox[2]{#2}
  \newcommand*\fsize{\dimexpr\f@size pt\relax}
  \newcommand*\lineheight[1]{\fontsize{\fsize}{#1\fsize}\selectfont}
  \ifx\svgwidth\undefined
    \setlength{\unitlength}{226.77165354bp}
    \ifx\svgscale\undefined
      \relax
    \else
      \setlength{\unitlength}{\unitlength * \real{\svgscale}}
    \fi
  \else
    \setlength{\unitlength}{\svgwidth}
  \fi
  \global\let\svgwidth\undefined
  \global\let\svgscale\undefined
  \makeatother
  \begin{picture}(1,0.38204086)
    \lineheight{1}
    \setlength\tabcolsep{0pt}
    \put(0,0){\includegraphics[width=\unitlength,page=1]{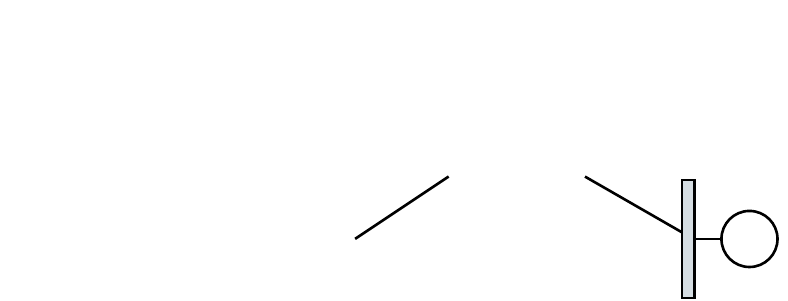}}
    \put(0.93147075,0.10129925){\color[rgb]{0,0,0}\makebox(0,0)[lt]{\begin{minipage}{0.61609064\unitlength}\raggedright $w'$\end{minipage}}}
    \put(0,0){\includegraphics[width=\unitlength,page=2]{mVAE_v3.pdf}}
    \put(0.75828859,0.1822195){\color[rgb]{0,0,0}\makebox(0,0)[lt]{\begin{minipage}{0.61609064\unitlength}\raggedright $z$\end{minipage}}}
    \put(0,0){\includegraphics[width=\unitlength,page=3]{mVAE_v3.pdf}}
    \put(0.93539952,0.28920413){\color[rgb]{0,0,0}\makebox(0,0)[lt]{\begin{minipage}{0.61609064\unitlength}\raggedright $x'$\end{minipage}}}
    \put(0,0){\includegraphics[width=\unitlength,page=4]{mVAE_v3.pdf}}
    \put(0.36045209,0.27597488){\color[rgb]{0,0,0}\makebox(0,0)[lt]{\begin{minipage}{0.61609064\unitlength}\raggedright $x$\end{minipage}}}
    \put(0,0){\includegraphics[width=\unitlength,page=5]{mVAE_v3.pdf}}
    \put(0.35652468,0.0881665){\color[rgb]{0,0,0}\makebox(0,0)[lt]{\begin{minipage}{0.61609064\unitlength}\raggedright $w$\end{minipage}}}
    \put(0,0){\includegraphics[width=\unitlength,page=6]{mVAE_v3.pdf}}
    \put(0.66175921,0.25097488){\color[rgb]{0,0,0}\makebox(0,0)[lt]{\begin{minipage}{0.61609064\unitlength}\raggedright $\mu$\end{minipage}}}
    \put(0.66175921,0.1131665){\color[rgb]{0,0,0}\makebox(0,0)[lt]{\begin{minipage}{0.61609064\unitlength}\raggedright $\sigma$\end{minipage}}}
    \put(0,0){\includegraphics[width=\unitlength,page=7]{mVAE_v3.pdf}}
    \put(0.245074,0.2825895){\color[rgb]{0,0,0}\makebox(0,0)[lt]{\begin{minipage}{0.61609064\unitlength}\raggedright $f_x$\end{minipage}}}
    \put(0,0){\includegraphics[width=\unitlength,page=8]{mVAE_v3.pdf}}
    \put(0.245074,0.09478113){\color[rgb]{0,0,0}\makebox(0,0)[lt]{\begin{minipage}{0.61609064\unitlength}\raggedright $f_w$\end{minipage}}}
    \put(0.04929651,0.37264125){\color[rgb]{0,0,0}\makebox(0,0)[lt]{\begin{minipage}{0.61609064\unitlength}\raggedright former detection\end{minipage}}}
    \put(0.76438437,0.296695){\color[rgb]{0,0,0}\makebox(0,0)[lt]{\begin{minipage}{0.61609064\unitlength}\raggedright $\tilde{z}$\end{minipage}}}
  \end{picture}
\endgroup

	\caption{Multi-modal VAE performing fusion of two sensor modalities camera and LiDAR by generating the missing $x$ from former $\tilde{z}$ in order to generate a new complete $z$.}
	\label{fig:sensor_fusion}
\end{figure}

Training the JMVAE on raw data resulted in non-sufficient results.
We conjecture that this effect is caused by the necessarily larger network architecture which causes the weights in deep layers to collapse and by the imbalance in the VAE's reconstruction loss term for varying modality dimensionality.
To derive a proper latent distribution, we therefore apply neuronal networks based scanline detectors $f_x$ and $f_w$ to derive feature vectors for each modality with dimensionality $D_x \equiv D_w$ (c.f. \autoref{fig:sensor_fusion}).

The experiment was performed on a comprehensive scenario with three classes (c.f. \autoref{tab:class}).
Modality $x$ is derived from a camera based feature extractor $f_x$ that distinguishes between red and green PoIs.
Modality $w$ is derived from a LiDAR based feature extractor $f_w$ that distinguishes between round and edgy PoIs.
The ability to unambiguously classify per modality is shown in \autoref{tab:class}.
We designed the JMVAE as depicted in \autoref{fig:sensor_fusion} with a Gaussian prior with unit variance, Gaussian variational distribution that is parametrized by the encoder network, and latent dimensionality of $D_z=2$ incorporating the sampled mean $\mu$ and variance $\sigma$ for each dimension.
All remaining layers are dense with 64 neurones and ReLU activation.

\begin{figure}
	\footnotesize
\begingroup
  \makeatletter
  \providecommand\color[2][]{
    \errmessage{(Inkscape) Color is used for the text in Inkscape, but the package 'color.sty' is not loaded}
    \renewcommand\color[2][]{}
  }
  \providecommand\transparent[1]{
    \errmessage{(Inkscape) Transparency is used (non-zero) for the text in Inkscape, but the package 'transparent.sty' is not loaded}
    \renewcommand\transparent[1]{}
  }
  \providecommand\rotatebox[2]{#2}
  \newcommand*\fsize{\dimexpr\f@size pt\relax}
  \newcommand*\lineheight[1]{\fontsize{\fsize}{#1\fsize}\selectfont}
  \ifx\svgwidth\undefined
    \setlength{\unitlength}{226.77165985bp}
    \ifx\svgscale\undefined
      \relax
    \else
      \setlength{\unitlength}{\unitlength * \real{\svgscale}}
    \fi
  \else
    \setlength{\unitlength}{\svgwidth}
  \fi
  \global\let\svgwidth\undefined
  \global\let\svgscale\undefined
  \makeatother
  \begin{picture}(1,0.59716456)
    \lineheight{1}
    \setlength\tabcolsep{0pt}
    \put(0,0){\includegraphics[width=\unitlength,page=1]{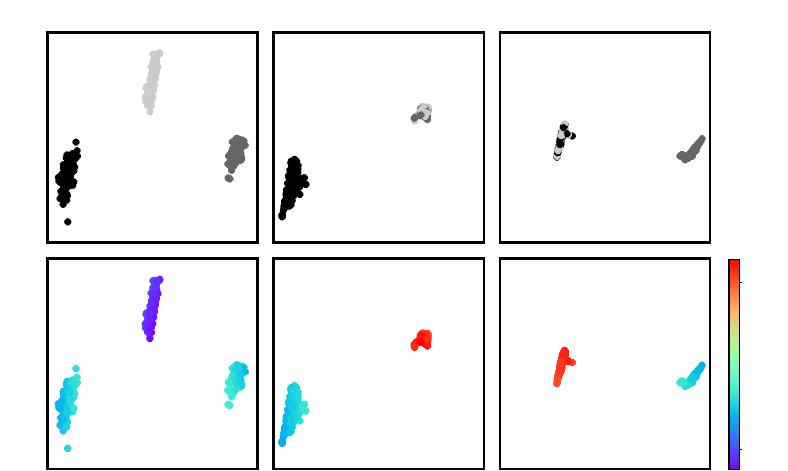}}
    \put(0.13863095,0.5992996){\color[rgb]{0,0,0}\makebox(0,0)[lt]{\begin{minipage}{0.61609064\unitlength}\raggedright $(x,w)$\end{minipage}}}
    \put(0.44294313,0.59929964){\color[rgb]{0,0,0}\makebox(0,0)[lt]{\begin{minipage}{0.61609067\unitlength}\raggedright $(x)$\end{minipage}}}
    \put(0.72617227,0.59929964){\color[rgb]{0,0,0}\makebox(0,0)[lt]{\begin{minipage}{0.61609066\unitlength}\raggedright $(w)$\end{minipage}}}
    \put(0.00680682,0.3884907){\color[rgb]{0,0,0}\rotatebox{90}{\makebox(0,0)[lt]{\begin{minipage}{0.61609066\unitlength}\raggedright class\end{minipage}}}}
    \put(0.00680682,0.0492287){\color[rgb]{0,0,0}\rotatebox{90}{\makebox(0,0)[lt]{\begin{minipage}{0.61609066\unitlength}\raggedright variance ($\sigma$)\end{minipage}}}}
    \put(0.9386273,0.03847047){\color[rgb]{0,0,0}\rotatebox{59.99999951}{\makebox(0,0)[lt]{\begin{minipage}{0.61609066\unitlength}\raggedright $1.05$\end{minipage}}}}
    \put(0.9386273,0.2505572){\color[rgb]{0,0,0}\rotatebox{59.99999951}{\makebox(0,0)[lt]{\begin{minipage}{0.61609066\unitlength}\raggedright $1.25$\end{minipage}}}}
    \put(0,0){\includegraphics[width=\unitlength,page=2]{classes_var.pdf}}
    \put(0.46464165,0.51978267){\color[rgb]{0,0,0}\makebox(0,0)[lt]{\begin{minipage}{0.61609064\unitlength}\raggedright (2) \& (3)\end{minipage}}}
    \put(0.64973569,0.38870873){\color[rgb]{0,0,0}\makebox(0,0)[lt]{\begin{minipage}{0.61609064\unitlength}\raggedright (1) \& (2)\end{minipage}}}
    \put(0.08655412,0.46356465){\color[rgb]{0,0,0}\makebox(0,0)[lt]{\begin{minipage}{0.61609064\unitlength}\raggedright (1)\end{minipage}}}
    \put(0.18534171,0.4453084){\color[rgb]{0,0,0}\makebox(0,0)[lt]{\begin{minipage}{0.61609064\unitlength}\raggedright (2)\end{minipage}}}
    \put(0.26675401,0.47176673){\color[rgb]{0,0,0}\makebox(0,0)[lt]{\begin{minipage}{0.61609064\unitlength}\raggedright (3)\end{minipage}}}
  \end{picture}
\endgroup

	\caption{Encoding of JMVAE showing classes and variances feeding in multi- and uni-modal detections.}
	\label{fig:classes_var}
\end{figure}

\autoref{fig:classes_var} shows, that the JMVAE is able to detect ambiguous classifications, as shown in \autoref{tab:class}.
Feeding in $(x,w)$ shows a clear separation of the three classes, while in the uni-modal cases, the distributions of ambiguous classes collapse to their mean.
Furthermore, the JMVAE drives up the variance due to the fact that the now collapsed distribution incorporates the old ones.
This property is absolutely reasonable as the KL-divergence attempts to find the best representative (i.e. the mean) and the reconstruction loss enforces the variance to extend to the non-collapsed classes during training.

The number of possible states of $\mathcal{M}\in\mathcal{S}$ is $\left|\mathcal{S}\right|=\#\text{PoI}!\cdot\#\text{PoI-encoding}^{\#\text{PoI} \#\text{mod.}}$, which is 384 for this relatively comprehensible experiment assuming only binary PoI-encodings.
Having continues encodings, as offered by the JMVAE, the task of controlling the robots by the means of handcrafted architectures based on the encodings becomes unfeasible.
Therefore, we train a DQN for which we select $z_n=\left(\left(\mu_{1,n},\sigma_{1,n}\right), \ldots,\left(\mu_{D_{z},n}, \sigma_{D_{z},n}\right)\right)\in \mathcal{M}$ and $I_n=1/\left\lVert \left( \sigma_{1,n}, \ldots, \sigma_{{D_{z}},n}\right)\right\rVert_2$.
The reward is shaped as follows: $r=\Delta I_n$ if the observation led to an increase of information; $r=-1$ if there is no increase in information; quitting the exploration by NOP always results in $r=0$.
The other crucial parameters (c.f. \cite{Mnih2015}) are $\gamma = .95$, $\epsilon_{\text{start}} = 1.$, $\epsilon_{\text{min}} = .01$, and $\epsilon_{\text{decay}} = .99$.
The Q-network has two dense layers with 24 neurones, ReLU activation, and a linear output layer trained by the Adam optimization algorithm ($\alpha = .001$).

\begin{figure}
	\footnotesize
	\includegraphics[width=1.0\textwidth]{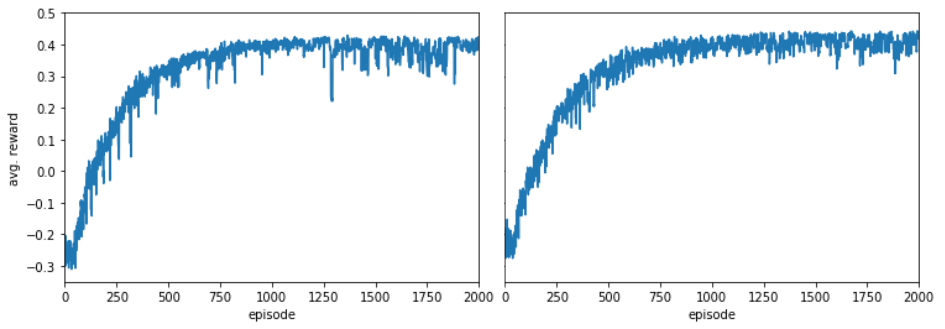}
	\caption{Average total reward over 2000 training epochs training a DQN for modality $x$ (left) and $w$ (right).}
	\label{fig:reward}
\end{figure}

We evaluated the training by the total reward the agent collects in an episode averaged over a 512 randomly sampled environments.
The average total reward metric is shown in \autoref{fig:reward}; it demonstrates the successful adaptation of each modality's network to our task (c.f. application video\footnote{\url{https://goo.gl/Edi92T}}).
\begin{table}
	\caption{Ability to unambiguously classify of camera ($x$) and LiDAR ($w$) detection for the experiment.}
	\footnotesize
	\begin{tabular}{ c c c c }
		\textbf{class vs. modality} & $x$ & $w$ & $x$ \& $w$\\
		\hline			
		green cylinder (1) & \checkmark &            & \checkmark\\
		red cylinder   (2) &            &            & \checkmark\\
		red cube       (3) &            & \checkmark & \checkmark
	\end{tabular}
	\label{tab:class}
\end{table}

\section{CONCLUSIONS AND FUTURE WORK}
\label{seq:conclusion}
A complete framework implementation has been proposed that is able to learn and control multiple robots with various sensor modalities in a comprehensive three-class example.
Further work will adapt VAEs to multiple classes and introduce a generic mapping approach.
\bibliographystyle{IEEEtran}
\bibliography{root}

\begin{thebibliography}{1}
\providecommand{\url}[1]{#1}
\csname url@rmstyle\endcsname
\providecommand{\newblock}{\relax}
\providecommand{\bibinfo}[2]{#2}
\providecommand\BIBentrySTDinterwordspacing{\spaceskip=0pt\relax}
\providecommand\BIBentryALTinterwordstretchfactor{4}
\providecommand\BIBentryALTinterwordspacing{\spaceskip=\fontdimen2\font plus
\BIBentryALTinterwordstretchfactor\fontdimen3\font minus
  \fontdimen4\font\relax}
\providecommand\BIBforeignlanguage[2]{{%
\expandafter\ifx\csname l@#1\endcsname\relax
\typeout{** WARNING: IEEEtran.bst: No hyphenation pattern has been}%
\typeout{** loaded for the language `#1'. Using the pattern for}%
\typeout{** the default language instead.}%
\else
\language=\csname l@#1\endcsname
\fi
#2}}

\bibitem{Tai2016}
\BIBentryALTinterwordspacing
L.~Tai, J.~Zhang, M.~Liu, J.~Boedecker, and W.~Burgard, ``{A Survey of Deep
  Network Solutions for Learning Control in Robotics: From Reinforcement to
  Imitation},'' vol.~14, no.~8, pp. 1--19, 2016.
\BIBentrySTDinterwordspacing

\bibitem{Mnih2013}
\BIBentryALTinterwordspacing
V.~Mnih, K.~Kavukcuoglu, D.~Silver, A.~Graves, I.~Antonoglou, D.~Wierstra, and
  M.~Riedmiller, ``{Playing Atari with Deep Reinforcement Learning},'' pp.
  1--9, 2013.
\BIBentrySTDinterwordspacing

\bibitem{Nowe2012}
\BIBentryALTinterwordspacing
A.~Nowe, P.~Vrancx, and Y.-M.~D. Hauwere, ``{Game Theory and Multi-agent
  Reinforcement Learning},'' 2012, vol.~12, no. January.
\BIBentrySTDinterwordspacing

\bibitem{Higgins2017}
\BIBentryALTinterwordspacing
I.~Higgins, A.~Pal, A.~A. Rusu, L.~Matthey, C.~P. Burgess, A.~Pritzel,
  M.~Botvinick, C.~Blundell, and A.~Lerchner, ``{DARLA: Improving Zero-Shot
  Transfer in Reinforcement Learning},'' 2017.
\BIBentrySTDinterwordspacing

\bibitem{Kingma2014}
\BIBentryALTinterwordspacing
D.~P. Kingma, D.~J. Rezende, S.~Mohamed, and M.~Welling, ``{Semi-Supervised
  Learning with Deep Generative Models},'' 2014.
\BIBentrySTDinterwordspacing

\bibitem{Suzuki2017}
M.~Suzuki, K.~Nakayama, and Y.~Matsuo, ``{JOINT MULTIMODAL LEARNING WITH DEEP
  GENERATIVE MODELS},'' pp. 1--12, 2017.

\bibitem{Herbrechtsmeier2016a}
S.~Herbrechtsmeier, T.~Korthals, T.~Sch{\"{o}}pping, and U.~R{\"{u}}ckert,
  ``{AMiRo: A modular {\&} customizable open-source mini robot platform},''
  2016.

\bibitem{Mnih2015}
\BIBentryALTinterwordspacing
V.~Mnih, K.~Kavukcuoglu, D.~Silver, A.~A. Rusu, J.~Veness, M.~G. Bellemare,
  A.~Graves, M.~Riedmiller, A.~K. Fidjeland, G.~Ostrovski, S.~Petersen,
  C.~Beattie, A.~Sadik, I.~Antonoglou, H.~King, D.~Kumaran, D.~Wierstra,
  S.~Legg, and D.~Hassabis, ``{Human-level control through deep reinforcement
  learning},'' vol. 518, no. 7540, pp. 529--533, 2015.
\BIBentrySTDinterwordspacing

\end{thebibliography}

\end{document}